 \documentclass[pmlr,twocolumn]{jmlr} 

\usepackage{booktabs}
\usepackage[load-configurations=version-1]{siunitx} 

\theorembodyfont{\upshape}
\theoremheaderfont{\scshape}
\theorempostheader{:}
\theoremsep{\newline}

\jmlrvolume{ML4H Extended Abstract Arxiv Index}
\jmlryear{2020}
\jmlrsubmitted{2020}
\jmlrpublished{}
\jmlrworkshop{Machine Learning for Health (ML4H) 2020} 

\title[Cost-Sensitive Classification for Tuberculosis Verbal Screening]{Cost-Sensitive Machine Learning Classification for Mass Tuberculosis Verbal Screening}

\author{%
\Name{Ali Akbar Septiandri} \Email{ali.septiandri@gmail.com}\\
\Name{Aditiawarman} \Email{adit@inovasisehat.co.id}\\
\Name{Roy Tjiong} \Email{rtjiong@gmail.com}\\
\addr Inovasi Sehat Indonesia, Jakarta, Indonesia
\AND
\Name{Erlina Burhan} \Email{erlina\_burhan@yahoo.com}\\
\addr University of Indonesia, Jakarta, Indonesia
\AND
\Name{Anuraj Shankar} \Email{anuraj.shankar@ndm.ox.ac.uk}\\
\addr Oxford University, Oxford, UK
}

\begin{document}

\maketitle

\begin{abstract}
Score-based algorithms for tuberculosis (TB) verbal screening perform poorly, causing misclassification that leads to missed cases and unnecessary costly laboratory tests for false positives. We compared score-based classification defined by clinicians to machine learning classification such as SVM-RBF, logistic regression, and XGBoost. We restricted our analyses to data from adults, the population most affected by TB, and investigated the difference between untuned and unweighted classifiers to the cost-sensitive ones. Predictions were compared with the corresponding GeneXpert\textregistered MTB/Rif results. After adjusting the weight of the positive class to 40 for XGBoost, we achieved 96.64\% sensitivity and 35.06\% specificity. As such, the sensitivity of our identifier increased by 1.26\% while specificity increased by 13.19\% in absolute value compared to the traditional score-based method defined by our clinicians. Our approach further demonstrated that only 2000 data points were sufficient to enable the model to converge. The results indicate that even with limited data we can actually devise a better method to identify TB suspects from verbal screening.
\end{abstract}

\section{Introduction}

In developing countries, affording state-of-the-art technology such as using polymerase chain reaction (PCR) devices to detect tuberculosis (TB) cases, e.g. GeneXpert\textregistered MTB/Rif, is still hard. Yet, these countries are the ones where the number of TB cases is still high. According to the Global Tuberculosis Report 2016 \citep{world2016global}, ``Six countries accounted for 60\% of the new cases: India, Indonesia, China, Nigeria, Pakistan and South Africa.'' Recent cases of COVID-19 have shown that population-wide testing and contact tracing are necessary for deadly communicable diseases \citep{panovska2020determining, matukas2020aggressively}. Mass verbal screening can help to overcome this challenge in early detection of tuberculosis \citep{bothamley2002screening, malik2018improving}. However, this solution tends to have low specificity. This work addresses the problem by developing a machine learning model to increase the verbal screening performance.

Most medical diagnosis cases have to deal with the imbalanced dataset problem \citep{jothi2015data}. Since the majority of the people are healthy, the classification results are more likely to be negative. Since we want to prevent an epidemic because TB is a communicable disease, our goal is then to keep the sensitivity as high as possible while trying to increase the specificity.

We utilised several machine learning algorithms, namely support vector machines (SVM) with Gaussian (RBF) kernel \citep{fernandez2014we} and XGBoost \citep{Chen2016xgb}. As a benchmark, we also employed a logistic regression model. We evaluated our models by comparing our predictions with the corresponding GeneXpert\textregistered MTB/Rif results. We used two score-based classification methods as the baselines in this study.

\section{Related Work}

Automated tuberculosis (TB) detection has been studied for nearly two decades. One way to detect TB cases automatically is by using a rule-based approach from both clinical and laboratory criteria \citep{autotb}. In a more recent approach, \cite{shamshirband2014tuberculosis} were able to achieve 87.00\% sensitivity and 86.12\% specificity in TB diagnosis using Artificial Immune Recognition System (AIRS) learned from supporting lab results. It is also shown in \citep{seixas2013artificial} that using artificial neural networks (ANN) could help them to reach more 94.5\% sensitivity and 91.0\% specificity in diagnosing pleural TB based on anamnesis variables and HIV status.

Although a real-time screening alert has been done in \citep{weng2011real}, machine learning was not employed in that study. Therefore, we proposed to do so in this paper aiming for a simpler but more effective tool to identify TB suspects compared to simple score or rule-based methods.

\section{Dataset}

\paragraph{Cohort}

The dataset in this study was collected by field workers. The data were collected from 12 November 2013 to 1 July 2016 from several sites in Jakarta, Indonesia. The dataset we are using in this experiment consists of 8732 rows where each row corresponds to an individual with 7508 TB- and 1224 TB+ from the screening process with their available GeneXpert\textregistered MTB/Rif result. We are also focusing on people with no history of TB before our study. Due to inaccuracy, we chose to remove the rows where the height is either unavailable or less than 120 cm.

We divided our dataset into three sets, i.e. training, validation, and test sets (60:20:20). We built our models using machine learning algorithms from the training set and then evaluated them with the validation set. Eventually, we tested the best models from each algorithm on the test set.


\paragraph{Features}

The features in this study are symptoms, comorbidities, and characteristics commonly found in TB patients. Four attributes in this dataset are of numerical values, i.e. height, weight, age, and cough duration (in days). We also used the sex of the people screened as the attributes for machine learning algorithms. The statistical properties of the features can be seen in Table~\ref{tab:dataset}.

Aside from derived attributes like the body mass index (BMI), the attributes were acquired by verbally asking the subjects. We only marked `Yes' as the value of diabetes, kidney failure, asthma, COPD, and HIV questions if the subjects had done the appropriate lab test before. We can see that breathing shortness, chest pain, diabetes, family diabetes, TB exposure, kidney failure, asthma, COPD, and HIV contain lots of unknown values. This is because they were added later in the field study. Comorbidity in particular needs to be confirmed by a lab test for our screeners to report them as a yes/no.

\begin{table}[htbp]
\footnotesize
\floatconts
  {tab:dataset}
  {\caption{Statistical Properties of the Features}}
  {\begin{tabular}{lccc}
  \toprule
  \bfseries Feature & \bfseries Yes & \bfseries No & \bfseries Unknown \\
  \midrule
  cough & 5480 & 3242 & 10 \\
  fever & 1777 & 6955 & 0 \\
  haemoptysis & 646 & 7758 & 0 \\
  night sweat & 1365 & 7367 & 0 \\
  weight loss & 1890 & 6842 & 0 \\
  fatigue & 2185 & 6547 & 0 \\
  breathing shortness & 189 & 1370 & 7173 \\
  chest pain & 123 & 1436 & 7173 \\
  diabetes & 969 & 5092 & 2671 \\
  family diabetes & 137 & 468 & 8127 \\
  TB exposure & 1414 & 4808 & 2510 \\
  kidney failure & 75 & 4880 & 3777 \\
  asthma & 717 & 4831 & 3184 \\
  COPD & 232 & 4710 & 3790 \\
  HIV & 251 & 4695 & 3786 \\
  active smoker & 4313 & 4419 & 0 \\
  \bottomrule
  \end{tabular}}
\end{table}

\section{Methods}


\paragraph{Classification Algorithms}

Our clinicians devised two scoring methods that were used throughout the project. When the total score of a person exceeds a certain threshold, we classify that person as a TB suspect. Those TB suspects were then tested by GeneXpert\textregistered MTB/Rif to confirm our finding. The pseudo-code for these methods can be seen in Algorithm~\ref{alg:scoring1} and Algorithm~\ref{alg:scoring2} in the Appendix. In these two algorithms, a person will be categorised as \emph{underweight} if their BMI is lower than 18.5.


This method needs experts to define the scoring scheme, i.e. how each attribute contributes to our suspicion about whether a person is TB+. To deal with this problem, we proposed a machine learning classification which continues to learn as the number of data grows larger. We employed support vector machine (SVM) \citep{boser1992training}, XGBoost \citep{Chen2016xgb}, and logistic regression models to classify the individuals.

\paragraph{Evaluation Approach/Study Design}

We trained the untuned version of each machine learning algorithm first to get an idea of how a cost-sensitive approach would later improve our models. We kept the parameters to the default settings and imputed the unknown/null values to the most frequent values. After that, we trained the models using a different weighting of positive class with the chosen hyperparameters. We tuned the hyperparameters with several different class weights, such as 10, 15, 20, 25, 30, 35, 40, 45, and 50. We did a grid search with 3-fold cross-validation using our training data to get the optimal AUC score. We chose AUC score to optimise both sensitivity and specificity.


\section{Results}


\paragraph{Untuned Classifiers}

The untuned classifiers tend to favour the negative class as can be seen in Table~\ref{tab:early-performance}. The machine learning classifiers performed worse than the score-based classifiers because the training set is highly imbalanced while the score-based classifiers are optimised for sensitivity.

\begin{table*}[htbp]
\footnotesize
\floatconts
  {tab:early-performance}
  {\caption{Classifier Performance (in \%)}}
  {\begin{tabular}{lcccccc}
  \toprule
  & \multicolumn{2}{c}{\bfseries Untuned} & \multicolumn{2}{c}{\bfseries Subsampling} & \multicolumn{2}{c}{\bfseries Test Set (Cost-sensitive)} \\
  \bfseries Classifier & \bfseries Sensitivity & \bfseries Specificity & \bfseries Sensitivity & \bfseries Specificity & \bfseries Sensitivity & \bfseries Specificity \\
  \midrule
  LogReg  & 16.24 & 97.49  &      68.80      & \textbf{74.74} & \textbf{98.32} & 22.80          \\
  XGBoost & 23.93 & 97.49  &      68.80      &     74.27      & 96.64          & 35.06          \\
  SVM-RBF &  0.00 & 100.00 &  \textbf{73.50} &     71.03      & 96.64          & 29.82          \\
  Score-1 & 93.16 & 20.64  & -               & -              & 95.38          & 21.87          \\
  Score-2 & 84.19 & 46.30  & -               & -              & 88.24          & \textbf{43.94} \\
  \bottomrule
  \end{tabular}}
\end{table*}

\paragraph{Cost-sensitive Classifiers}


In SVM case, we got higher sensitivity than Score-2 and higher specificity than Score-1 when we used at least 15 as the positive class weight. While the class weight is less than 20, we can also produce at least the same specificity as Score-1 with SVM. With XGBoost, we got an even better result: increasing the \textbf{specificity} from 20.63\% to \textbf{35.78\%} while only reducing the \textbf{sensitivity} from 93.16\% to \textbf{92.74\%} based on Score-1's result (class weight: 40). Yet, this specificity is still 10.45\% away from Score-2 specificity (46.23\%). Logistic regression also produces a quite decent result when using 20 as the class weight, i.e. 93.16\% sensitivity and 34.26\% specificity.


\paragraph{The Effect of Dataset Size}

\begin{figure}[htbp]
\floatconts
  {fig:dataset}
  {\caption{Dataset Size Effect}}
  {\includegraphics[width=\linewidth]{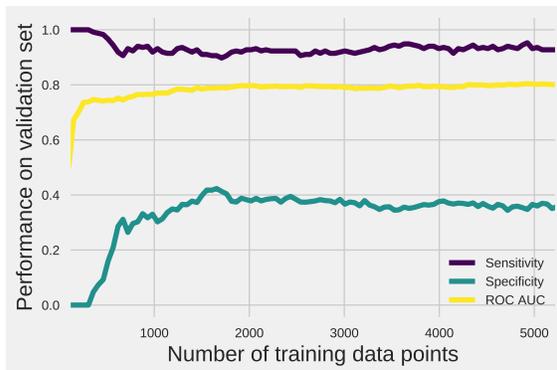}}
\end{figure}

As shown in Figure~\ref{fig:dataset}, all the metrics seem to converge when the number of training data is $\approx 2000$. Our exploration with the subsampling method also turned out to be ineffective as shown in Table~\ref{tab:early-performance}. These results imply that using cost-sensitive classifiers is preferred to using subsampling to overcome the imbalanced class problem.

\paragraph{Evaluation on Test Set}


Although the best specificity is achieved by the Score-2 method, the sensitivity is low despite being used for early stage TB diagnostics. Our best compromise is to use the XGBoost model where the sensitivity and specificity are the second-best overall.

\section{Conclusions}

We were able to increase the specificity while still keeping the sensitivity relatively high by using machine learning algorithms. We showed that using imbalanced weighting for the classes is necessary for this task. Furthermore, the best model converged after around 2000 data points. Therefore, we might use this as a reference point in further experiments.

Increasing the specificity by more than 10\% as achieved in this study means saving lots of money and time to do unnecessary tests. We can achieve this without losing any sick people in the screening process. In a more practical sense, the best model from our work can be installed on smartphones where verbal screeners can rapidly identify TB suspects, especially in rural and poor areas where access to clinics might be limited. The suspects can then be taken to better health facilities to be tested further to confirm the findings.

\acks{We thank Stop TB Partnership for providing GeneXpert\textregistered MTB/Rif instruments through the TB REACH Wave 3 program.}

\bibliography{jmlr}

\appendix

\begin{algorithm2e*}[htbp]
\caption{Score-1}
\label{alg:scoring1}
$score \leftarrow 0$\;
\For{$s \in \{cough, fever, night\_sweat, weight\_loss, fatigue\}$}{
  \If{$s = true$}{
    $score \longleftarrow score+1$\;
  }
}

\If{$hasKidneyFailure || hasAsthma || hasCOPD || hasHIV$}{
  $score \longleftarrow score+1$\;
}

\If{$hasTBExposure$}{
  $score \longleftarrow score+1$\;
}

\If{$hasDiabetes$}{
  $score \longleftarrow score+1$\;
}

\If{$isUnderweight$}{
  $score \longleftarrow score+1$\;
}

\If{$isActiveSmoker$}{
  $score \longleftarrow score+1$\;
}

\If{$coughDuration\ge14$}{
  $score \longleftarrow score+2$\;
}

\If{$hasHaemoptysis$}{
  $score \longleftarrow score+3$\;
}

\textbf{return} $score \ge 2$\;
\end{algorithm2e*}

\begin{algorithm2e*}
\caption{Score-2}
\label{alg:scoring2}
$score \leftarrow 0$\;
\For{$s \in \{cough, fever, night\_sweat, weight\_loss, fatigue, breathing\_shortness, chest\_pain\}$}{
  \If{$s = true$}{
    $score \leftarrow score+2$\;
  }
}

\If{$hasKidneyFailure || hasAsthma || hasCOPD || hasHIV$}{
  $score \leftarrow score+1$\;
}

\If{$hasTBExposure$}{
  $score \leftarrow score+1$\;
}

\If{$hasDiabetes$}{
  $score \leftarrow score+1$\;
}

\If{$isUnderweight$}{
  $score \leftarrow score+1$\;
}

\If{$isActiveSmoker$}{
  $score \leftarrow score+1$\;
}

\If{$coughDuration\ge14$}{
  $score \leftarrow score+4$\;
}

\If{$hasHaemoptysis$}{
  $score \leftarrow score+4$\;
}

\textbf{return} $score \ge 4$
\end{algorithm2e*}

\end{document}